\documentclass[10pt,twocolumn,letterpaper]{article}

\usepackage[pagenumbers]{cvpr} 

\usepackage{microtype}

\renewcommand{\paragraph}[1]{\vspace{0.2em}\noindent\textbf{#1}}

\definecolor{cvprblue}{rgb}{0.21,0.49,0.74}
\usepackage[pagebackref,breaklinks,colorlinks,allcolors=cvprblue]{hyperref}
\usepackage[accsupp]{axessibility}

\def\paperID{} 
\def\confName{CVPR}
\def\confYear{2026}

\title{StreamAvatar: Streaming Diffusion Models for Real-Time Interactive Human Avatars}

\author{
Zhiyao Sun$^{1}$\footnotemark[1] \footnotemark[2] \quad
Ziqiao Peng$^{2}$\footnotemark[1] \quad
Yifeng Ma$^{3}$\footnotemark[1] \quad
Yi Chen$^{3}$ \quad
Zhengguang Zhou$^{3}$ \quad
Zixiang Zhou$^{3}$ \\
Guozhen Zhang$^{4}$ \quad
Youliang Zhang$^{1}$ \quad
Yuan Zhou$^{3}$\footnotemark[3] \quad
Qinglin Lu$^{3}$\footnotemark[4] \quad
Yong-Jin Liu$^{1}$\footnotemark[4]
\vspace{5pt} \\
$^{1}$Tsinghua University \quad
$^{2}$Renmin University of China \quad
$^{3}$Tencent Hunyuan\quad
$^{4}$Nanjing University
\vspace{5pt} \\
{\tt\small Project Page: \url{https://streamavatar.github.io}}
}

\begin{document}

\twocolumn[{%
\renewcommand\twocolumn[1][]{#1}%
\maketitle
\vspace{-8mm}
\includegraphics[width=\linewidth]{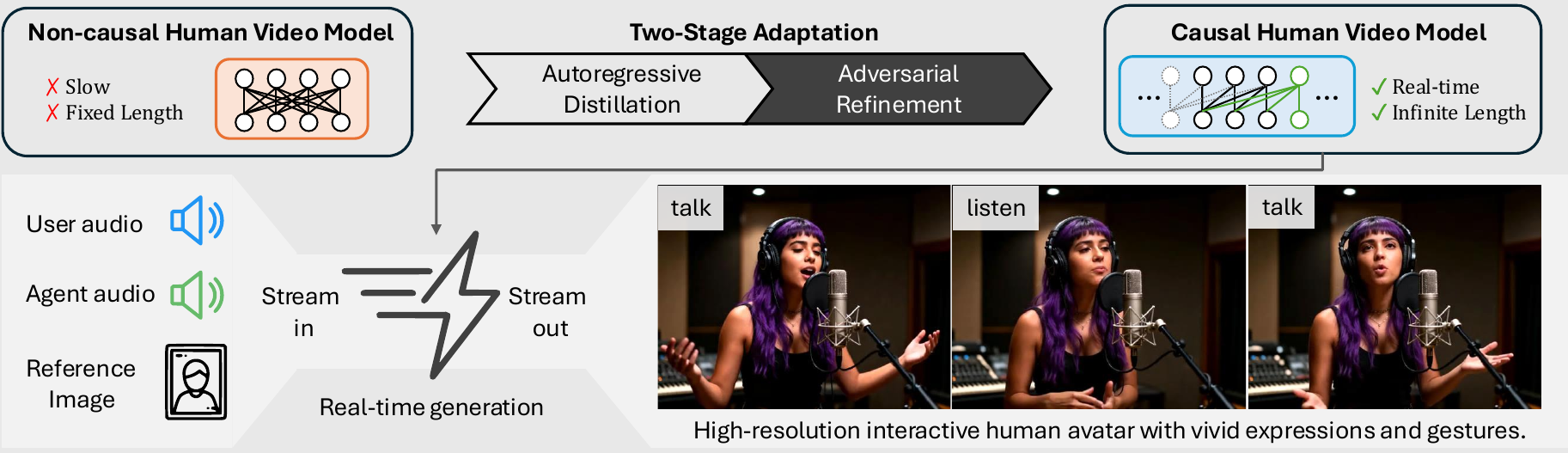}
\vspace{-5mm}
\captionof{figure}{We propose StreamAvatar, which adapts human video diffusion models for real-time, streaming, and interactive video generation through a two-stage framework. Given a reference image and user/agent audio streams as input, StreamAvatar generates high-resolution streaming human video in real-time, producing vivid talking/listening expressions and gestures.}
\vspace{1em}
\label{fig:teaser}
}]

\renewcommand{\thefootnote}{\fnsymbol{footnote}}
\footnotetext[1]{Equal contribution. \quad \textsuperscript{\ddag}Project leader. \quad \textsuperscript{\S}Corresponding author.}
\footnotetext[2]{Work done during internship at Tencent Hunyuan.}

\begin{abstract}
Real-time, streaming interactive avatars represent a critical yet challenging goal in digital human research. Although diffusion-based human avatar generation methods achieve remarkable success, their non-causal architecture and high computational costs make them unsuitable for streaming. Moreover, existing interactive approaches are typically restricted to the head-and-shoulder region, limiting their ability to produce gestures and body motions. To address these challenges, we propose a two-stage autoregressive adaptation and acceleration framework that applies autoregressive distillation and adversarial refinement to adapt a high-fidelity human video diffusion model for real-time, interactive streaming. To ensure long-term stability and consistency, we introduce three key components: a Reference Sink, a Reference-Anchored Positional Re-encoding (RAPR) strategy, and a Consistency-Aware Discriminator. Building on this framework, we develop a one-shot, interactive, human avatar model capable of generating both natural talking and listening behaviors with coherent gestures. Extensive experiments demonstrate that our method achieves state-of-the-art performance, surpassing existing approaches in generation quality, real-time efficiency, and interaction naturalness.
\end{abstract}    
\vspace{-3mm}
\section{Introduction}
\label{sec:intro}

Generating realistic, expressive, and versatile digital humans is a long-standing goal in computer vision and graphics, driving progress across a broad spectrum of research topics~\cite{SadTalker, EMO, Hallo3, Tailor, LDSwap, Zhang24Learning}. Among these, real-time interactive human avatars hold particular value, enabling fluid, dynamic communication in applications such as education, entertainment, and virtual assistants.
Recently, diffusion models have achieved remarkable success in human avatar generation~\cite{Hallo3, Sonic, FantasyTalking, OmniHuman-1, EchoMimicV3, HunyuanVideo-Avatar, OmniAvatar, StableAvatar, DiffPoseTalk}.
However, in real-time interactive settings, diffusion-based human avatar generation methods still face three key challenges:
    1) \emph{Real-time streaming:} Most methods are inherently unsuitable for real-time streaming. Firstly, the iterative denoising process and long-context attention incur a prohibitive amount of computation. Secondly, the non-causal bidirectional attention mechanism requires the entire video sequence to be processed at once, which is incompatible with streaming generation.
    2) \emph{Long-duration generation stability:} Streaming interaction requires generating long-duration videos; however, existing methods suffer from error accumulation and drifting when extended over time, resulting in degraded performance.
    3) \emph{Talking--listening interaction:} Current methods focus on \emph{one-way talking} generation while neglecting the avatar’s \emph{listening} state. In conversational scenarios, not modeling the listening state makes the interaction feel unnatural. Although some studies explore listening video generation~\cite{RLHG, L2L, CustomListener, DiffListener} or interactive video generation~\cite{ARIG, INFP}, they model only the facial or head area and fail to capture full-body expressiveness and reactivity.

To address these challenges, we propose StreamAvatar, which achieves real-time streaming interactive human video generation. 
StreamAvatar not only produces realistic results during speaking, but also generates rich and responsive listening motions conditioned on the interlocutor’s audio.
Specifically, we first train a powerful yet slow non-causal diffusion model for interactive human avatar generation as the teacher model, capable of producing both \emph{talking and listening} behaviors. 
To achieve such interaction capability, we utilize a talking-listening audio mask to extract distinct talking and listening audio features, which are then injected through dedicated talking/listening modules of the model. This design allows the model to generate expressive co-speech or listening expressions and gestures.

Then, we propose a two-stage autoregressive adaptation and acceleration framework for transforming this teacher model into a real-time, streaming student model. In the \emph{first} stage, we adapt the attention mechanism of the model from bidirectional attention into block-wise causal attention, and then perform distillation from the teacher model to obtain a few-step autoregressive generator, which significantly accelerates inference, reducing the DiT denoising process to 1/40 of its original runtime.
To solve the critical challenges of long-video stability and identity preservation, we also introduce novel mechanisms for attention: a \emph{Reference Sink}, which enforces persistent attention to the reference frame, and a \emph{Reference-Anchored Positional Re-encoding (RAPR)} strategy, which resolves the train--test mismatch and attention decay in long-sequence inference. In the \emph{second} stage, we propose an adversarial refinement process to resolve the quality degradation such as distortion and blur caused by distillation, and improve long video consistency. Specifically, we introduce a \emph{multitask consistency-aware discriminator}, which assesses both the realism of individual frames and the consistency over all generated frames. The adversarial refinement stage significantly improves the generation quality and stability of our causal generator.  

Our main contributions are summarized as follows:
\begin{itemize}
    \item We introduce a two-stage autoregressive adaptation and acceleration framework for human video diffusion models. The first stage performs autoregressive distillation to convert a bidirectional diffusion model into a real-time, streaming generator. The second stage performs adversarial refinement, which significantly improves generation quality and stability.
    \item To improve consistency and stability for long video generation, we propose three techniques, namely Reference Sink, Reference-Anchored Positional Re-encoding, and a consistency-aware discriminator. 
    \item Building on this framework, we develop StreamAvatar, a one-shot, real-time, streaming interactive human video model that can generate both talking and listening behaviors with coherent expressions, gestures, and transitions.
    \item Extensive experiments demonstrate that our method outperforms or performs competitively with existing state-of-the-art approaches, while operating significantly faster. 
\end{itemize}

\section{Related Work}
\label{sec:related}

\subsection{Audio-Driven Avatar Video Generation} \label{subsec:talking_avatar}
\paragraph{Traditional and 3D-Based Methods.} 
Early efforts rely on image-to-image translation \cite{Wav2Lip, PC-AVS} or image warping \cite{MakeItTalk} to generate lip-synced videos, but are often limited to the mouth region and fail to produce natural head motion. Later works~\cite{SadTalker, MusicFace, StyleTalk, VASA-1, Ma25Exploring} map audio to intermediate representations (e.g., 3DMMs or latent codes) and render videos with lightweight decoders, improving fidelity but still limited by the expressiveness of these intermediates. Another line of methods directly models the subject with audio-conditioned 3D representations such as Neural Radiance Fields~\cite{AD-NeRF, GeneFace, Ma25Decoupled} or 3D Gaussian Splatting~\cite{TalkingGaussian, SyncTalk++}. However, they typically require per-subject training, making them unable to generalize in a one-shot manner.

\paragraph{Diffusion-Based Avatars.} 
Recently, diffusion models have set a new standard for generation quality and one-shot generalization in the avatar space. A large body of work has focused on integrating audio features into large-scale video diffusion models to generate high-fidelity, expressive portraits~\cite{EMO, EchoMimic, Hallo3, Sonic, FantasyTalking, MagicTalk} and semi/full-body videos~\cite{CyberHost, OmniHuman-1, HunyuanVideo-Avatar, OmniAvatar, EchoMimicV2, EchoMimicV3, StableAvatar}. While these methods produce state-of-the-art visual quality, their reliance on an iterative, bidirectional denoising process makes them computationally prohibitive and fundamentally unsuitable for real-time, streaming applications. Our work bridges this gap, targeting the generation of high-fidelity human avatars in real-time.

\subsection{Streaming Video Diffusion Models} \label{subsec:streaming_video}
A large number of video diffusion models are limited to fixed-length generation due to their use of bidirectional attention. Common workarounds, such as motion or overlapping frames, often create unnatural transitions, identity drift, and cumulative errors. Moreover, these methods cannot reduce generation latency. To enable efficient, streaming generation, one prominent line of work focuses on distilling bidirectional models into few-step causal autoregressive systems. CausVid~\cite{CausVid} pioneered this by re-architecting a bidirectional DiT with block-causal attention, using Diffusion Forcing (DF)~\cite{DiffusionForcing} and distribution matching distillation (DMD)~\cite{DMD}. Self-Forcing~\cite{SelfForcing} later improved upon DF by identifying a critical train-test mismatch and introducing a ``student-forcing'' scheme, where the student model conditions on its own prior outputs. Seaweed-APT2~\cite{Seaweed-APT2} integrates adversarial post-training into autoregressive distillation to further boost generation quality. 

However, a key ``train-short, test-long mismatch'' challenge persists for these autoregressive methods: quality degrades when extrapolating to sequences far beyond the training horizon. LongLive~\cite{LongLive} and Self-Forcing++~\cite{Self-Forcing++} both address this by applying DMD to segments sampled from long videos generated by the student model. In contrast, we identify that this extrapolation failure also stems from an out-of-distribution issue induced by Rotary Positional Embeddings (RoPE). We propose a re-encoding mechanism to resolve this instability without the need of generating long videos. Similar to LongLive, we also employ an attention sink to mitigate identity drift.

\subsection{Interactive and Responsive Video Avatars} \label{subsec:interactive_avatar}
Natural human interaction is a dyadic process of speaking and listening. Recent avatar generation methods~\cite{LLIA, TalkingMachines} achieve real-time performance but only process talking audio, lacking temporally aligned listener responses. Conversely, listener-oriented approaches~\cite{RLHG, L2L, DIM, CustomListener, DiffListener} produce responses aligned with the interlocutor's audio yet do not model smooth speaking--listening transitions. Unified speaking–listening methods~\cite{INFP, ARIG} handle both states with natural transitions but confine motion to the head-and-shoulder region, precluding expressive hand and body gestures. MIDAS~\cite{MIDAS} unifies multimodal conditions into an LLM structure with a lightweight diffusion head for low-latency streaming, but requires per-identity finetuning and yields moderate visual quality. X-Streamer~\cite{X-Streamer} integrates speech understanding and audio-visual generation via a dual-Transformer architecture, yet suffers from limited motion diversity, stiff listening behavior, and low output resolution.

Our approach overcomes these limitations with one-shot, real-time, high-resolution generation that covers whole body motion and supports smooth speaking--listening transitions with synchronized listener responses.

\section{Method}

In this section, we first introduce our two-stage autoregressive adaptation and acceleration framework (\cref{subsec:method_distill}), and then develop a real-time, streaming interactive human video model (\cref{subsec:method_interact}) based on this framework.

\subsection{Streaming Human Video Diffusion Model} \label{subsec:method_distill}

\begin{figure*}[t]
  \centering
   \includegraphics[width=1\linewidth]{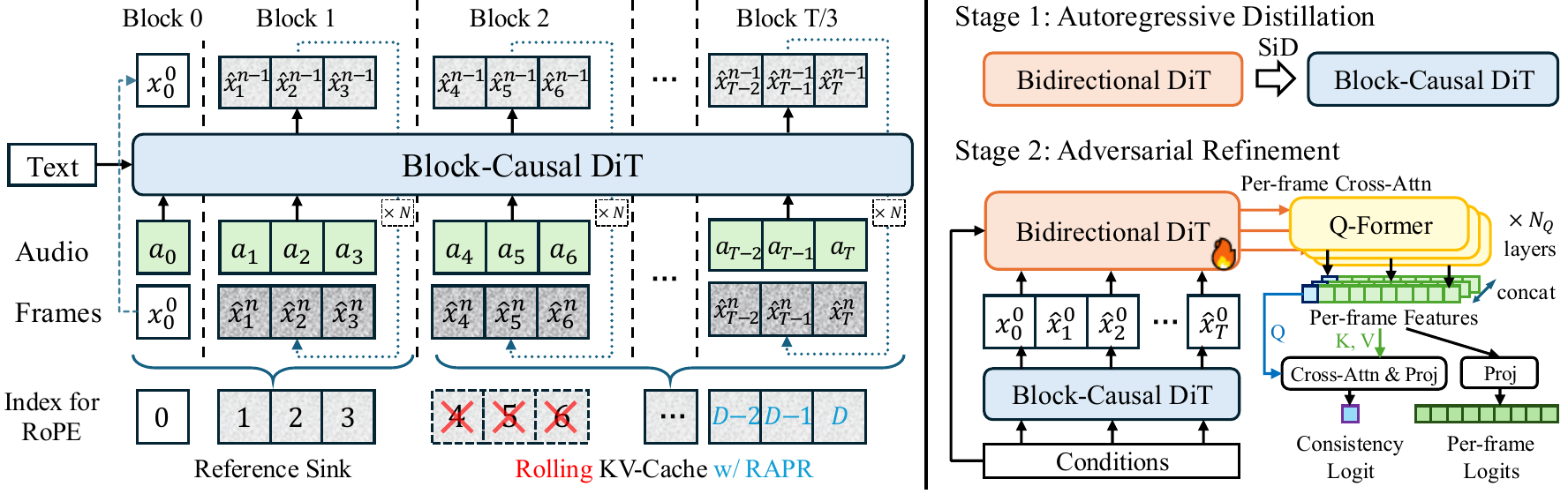}
   \caption{Overview of the two-stage autoregressive adaptation and acceleration framework. The original bidirectional DiT is first transformed into a block-causal DiT with block size $C=3$. Then, in stage 1, we apply Score Identity Distillation to distill from the bidirectional teacher into a block-causal student. A Reference Sink and Reference-Anchored Positional Re-encoding is introduced to improve long-term stability and consistency. In stage 2, we apply an adversarial refinement process guided by a Consistency-Aware Discriminator, to further improve generation quality, consistency, and stability.}
   \label{fig:ar_distill}
\end{figure*}

Recent human video generation approaches are generally built on video models constructed with diffusion transformers~\cite{DiT}. They typically adopt bidirectional attention, and take a reference image $x_0^0$, a text prompt $y$, and an audio feature sequence $\{a_t\}$ as input conditions. 
In the following, we describe our two-stage autoregressive adaptation and acceleration framework using this class of models as a representative example.

\subsubsection{Stage 1: Autoregressive Distillation.} This stage has two goals: (1) re-architect the model for autoregressive generation, and (2) distill its bidirectional, iterative denoising process into a causal, few-step model.

\paragraph{Re-architecture.} 
The distillation process operates on the Diffusion Transformer (DiT) within the VAE latent space. The original DiT is bidirectional over a fixed generation window $\{x_t^n\}_{t=0}^T$, where $n=0, 1, \dots, N$ denotes the diffusion timestep and $T+1$ is the window size. Inspired by prior works~\cite{CausVid, SelfForcing}, we split the generation window into smaller chunks: a clean 1-frame Reference chunk $\{x_0^0\}$ and subsequent Generation chunks $\{\{x_t^n\}_{t=s_i}^{e_i}\}_{i=1}^{T/C}$ of size $C$, where $s_i=(i-1) \cdot C+1$ and $e_i=i \cdot C$. See \cref{fig:ar_distill} for an example with $C=3$. We enforce causal attention between chunks and bidirectional attention within chunks, allowing the model to better capture local dynamics while enabling autoregression. To allow for efficient inference for long videos, a rolling KV cache is adopted to store a fixed window of context information. This modification preserves the original network weights, providing a strong starting point for distillation.

\paragraph{Distillation Pipeline.}
We first follow CausVid~\cite{CausVid} and introduce an initialization stage to stabilize subsequent distillation. Specifically, we use the teacher model to generate avatar videos from a small set of avatar images and audio clips and record the corresponding denoising trajectories to construct a dataset of Ordinary Differential Equation (ODE) solution pairs $(\{x_t^n\}, \{x_t^0\})$. Since our goal is to distill a \emph{few-step} autoregressive student generator, we retain only the timesteps used by the student model and train it to predict $\{x_t^0\}$ from $\{x_t^n\}$ for each chunk using a regression loss.
Next, we apply Score Identity Distillation (SiD)~\cite{SiD} to train the student model by distilling from the teacher. Importantly, we adopt the student-forcing scheme from Self Forcing~\cite{SelfForcing}, where the student model predicts the next chunk $\{\hat x_t\}_{t=s_j}^{e_j}$ conditioned on its own output from the previous chunks $\{\{\hat x_t\}_{t=s_i}^{e_i}\}_{i<j}$, thereby mitigating the training–test mismatch. Note that in the original Self Forcing, the next chunk’s denoising process is conditioned on clean (fully denoised) previous chunks $\{\hat x_t^0\}$, which requires an additional forward pass to update the KV-cache after denoising. However, we empirically find that omitting this update step does not noticeably degrade generation quality. This means the next chunk can instead be conditioned on noisy previous chunks $\{\hat x_t^1\}$, which reduces one forward pass per chunk and improves efficiency.

\paragraph{Reference Sink.} 
To avoid the context window growing infinitely, prior works~\cite{CausVid, SelfForcing} introduce a rolling KV cache. However, as new chunks are generated, the KV pairs for the reference frame are eventually evicted, leading to severe identity drift in long videos. We introduce an attention sink in the KV cache, where the KV pairs of the reference frame $x_{0}^{0}$ are permanently retained and never evicted. This ensures the model can always attend to the original identity. We also find that extending the sink to include the first generated chunk $\{\hat{x}_{t}^{1}\}_{t=s_{1}}^{e_{1}}$ further improves consistency.

\begin{figure}[t]
  \centering
   \includegraphics[width=\linewidth]{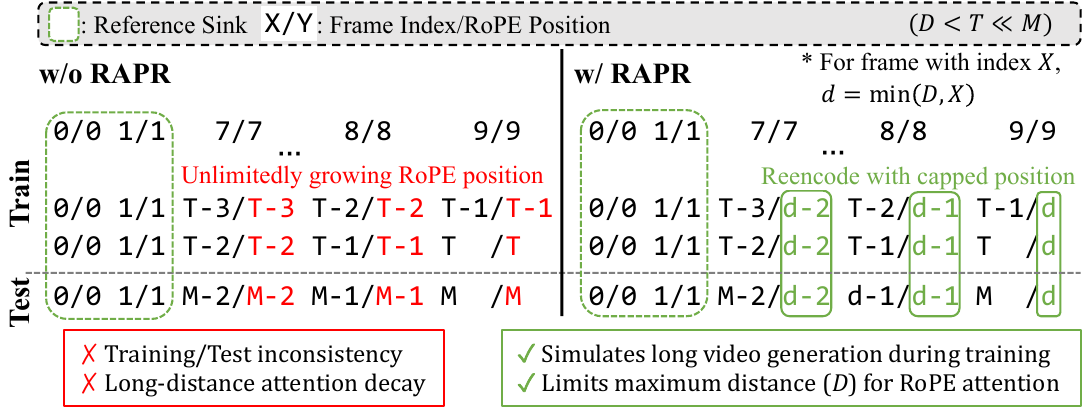}
   \caption{Vanilla RoPE vs RoPE with Reference-Anchored Positional Re-encoding (RAPR). RAPR improves long video generation without the need for long video training.}
   \label{fig:attention}
\end{figure}

\paragraph{Reference-Anchored Positional Re-encoding.} 
However, adding Reference Sink alone is insufficient for maintaining consistency in long video generation. We identify that this limitation primarily arises from two critical issues induced by the standard Rotary Positional Embedding (RoPE) \cite{RoPE}: 1) \emph{Train--Test Mismatch}: Vanilla RoPE assigns global frame indices as positional indices. As the model is trained only on short clips (e.g., $T$ frames), it never encounters the large positional indices required for long-duration streaming, leading to catastrophic out-of-distribution issues. 2) \emph{Attention Decay}: The long-term decay property inherent in RoPE causes the attention scores toward the Reference Sink to diminish as the generation window moves further away, exacerbating identity drift. To address this, we propose \emph{Reference-Anchored Positional Re-encoding (RAPR)} (See \cref{fig:attention}). RAPR changes how positional indices are managed in the KV cache. The mechanism is as follows: 1) We store \emph{non-encoded} keys in the KV cache. 2) When generating the current frame $x_t$, we calculate its capped distance to the reference $x_0$ with a maximum limit $D$ (i.e., $\min(t, D)$), which serves as its RoPE index. 3) We synchronously shift the indices of all other cached keys to maintain their correct relative positions based on this capped distance. RoPE is then applied to these re-calculated positions. RAPR provides two crucial benefits. First, by capping the maximum distance $D$, it prevents attention decay for distant frames, ensuring the model consistently attends to the reference. Second, it mitigates the train--test mismatch problem. By enabling RAPR during both training and inference, the model learns to operate within a finite positional-encoding space (defined by $D < T$). This allows the model to simulate long-video positional shifts during training using only short clips, improving stability during extended inference.

\subsubsection{Stage 2: Adversarial Refinement}
After the Stage 1 distillation, we obtain a few-step causal generator capable of real-time, streaming generation. However, due to the aggressive reduction in diffusion steps and architectural modifications, the distilled model may exhibit visual artifacts (e.g., blurring in hands or teeth) and temporal inconsistencies. To address these issues, we introduce an adversarial refinement stage featuring a novel \emph{consistency-aware} discriminator (see \cref{fig:ar_distill}). We first follow prior works~\cite{Seaweed-APT} to initialize the discriminator from the pretrained teacher model's backbone, and insert $N_Q$ Querying Transformers (Q-Formers)~\cite{BLIP-2} with learnable \emph{per-frame} queries into the intermediate layers to extract deep features for each frame. The discriminator’s logit output adopts a dual-branch design: 1) \emph{Local Realism Branch:} applies a linear projection directly to the per-frame features, producing per-frame logits that assess the realism of individual generated frames; 2) \emph{Global Consistency Branch:} enforces identity and temporal consistency by applying cross-attention between the reference frame’s feature and those of all subsequent frames, followed by a linear projection to yield a single logit. This global branch explicitly penalizes undesired deviation from the reference. We train the student generator against this discriminator using a relativistic adversarial loss~\cite{RelativisticLoss} and R1/R2 gradient penalty~\cite{Mescheder18Which}, as proposed in~\cite{R3GAN, Seaweed-APT}. Note that different from the distillation stage, the adversarial stage is trained with real video data, which directly optimize the generation distribution towards real distribution.

\subsection{Interactive Human Generation} \label{subsec:method_interact}
\begin{figure}[t]
  \centering
   \includegraphics[width=\linewidth]{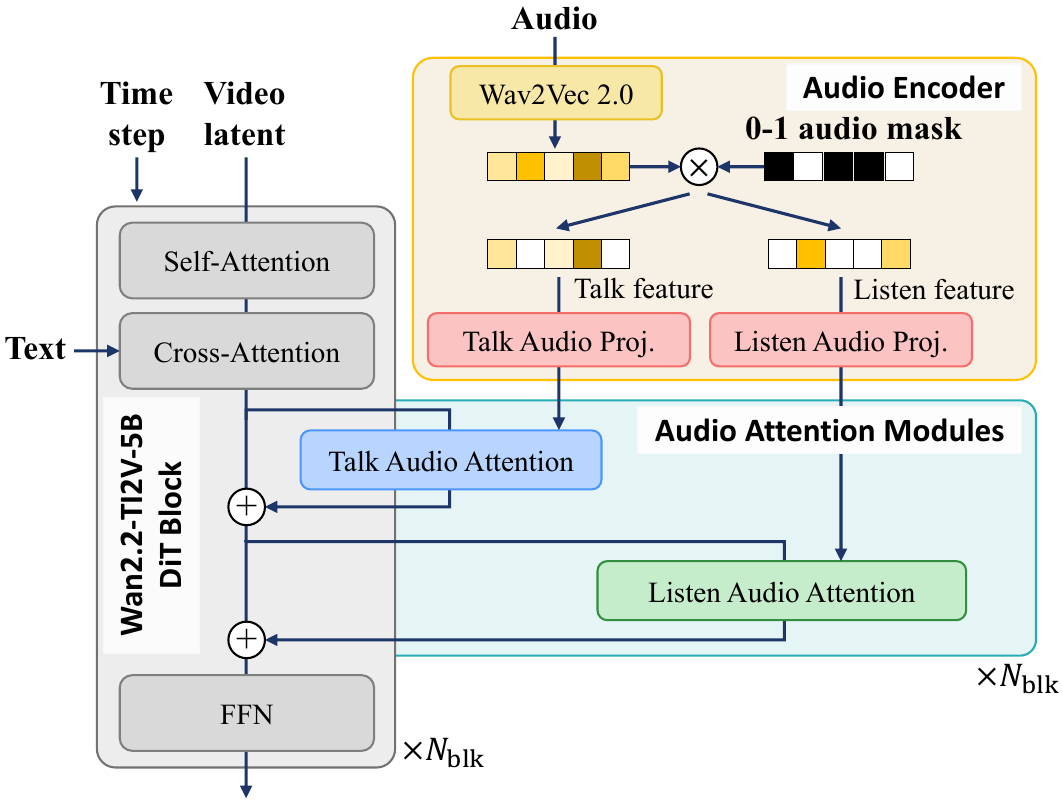}
   \caption{The architecture of our interactive human generation model. We extend the original video model with audio-related modules to support talking and listening audio conditioning.}
   \label{fig:structure}
\end{figure}

Our aim is to develop a real-time, streaming human video model that supports \emph{natural interactive human behaviors} in realistic scenes. We first train a bidirectional teacher model, and then distill it into a real-time causal model using our proposed framework. To start with, we adopt the Wan2.2-TI2V-5B~\cite{Wan} with $N_\text{blk}=30$ DiT blocks as the backbone. Then, we propose an audio encoder which takes a raw audio clip and an audio mask as input and produce separated talking and listening features. We also extend the DiT block with two audio attention modules to inject the audio features. \cref{fig:structure} shows the extended model architecture.

\paragraph{Audio Mask for Interaction Phase Identification.}
To distinguish between speaking and listening phases, we adopt an \emph{audio mask} rather than \emph{audio track separation} used in previous methods~\cite{INFP}.  
The audio mask is obtained through TalkNet~\cite{talknet}, a joint audio–video detection method. We observed that TalkNet provides more accurate temporal segmentation than audio separation methods. 
Audio track separation modifies waveforms, often producing signals that deviate from the data distribution used in Wav2Vec 2.0~\cite{Wav2Vec2} pretraining.  
Such deviations degrade the quality of extracted Wav2Vec features.
The audio mask avoids this issue by preserving the original waveform while marking each frame as either speaking ($1$) or listening ($0$).
It is applied \emph{after} Wav2Vec feature extraction, where it multiplicatively modulates the extracted representations rather than altering the raw waveform.
This design maintains high-fidelity audio features while providing precise temporal control over speaking and listening intervals.

\paragraph{Integration within the Generation Framework.}
The masked Wav2Vec features are injected into the video latent through two audio attention modules: 1) \emph{Talk Audio Attention}, which introduces talking cues to drive expressive human motion during speaking segments. 2) \emph{Listen Audio Attention}, which introduces listening cues to generate natural reactive behaviors during listening intervals.
All other layers in the framework, such as self-attention and cross-attention, remain audio-independent and focus on modeling visual and contextual dependencies. Note that our method disables text-based control and instead allows the model to generate natural motions purely from the audio. Specifically, we fix the text prompt to ``\emph{a person is speaking and listening.}''

\paragraph{Listen-and-Talk Generation Ability.}
By combining the base video model’s expressive generative capacity with phase-specific audio conditioning enabled by the audio mask, the system achieves \emph{natural interactive human generation}.
It produces smooth transitions between speaking and listening states and generates appropriate gestures and reactions, demonstrating realism and coherence far beyond traditional talk-only approaches. For additional details of this teacher model, please refer to Appendix \cref{ap_sec:teacher}.
\section{Experiment}

\subsection{Experimental Setups}

\paragraph{Dataset.}
Our data comes from a combination of SpeakerVid-5M~\cite{SpeakerVid-5M} and self-collected videos. We sample and filter $\sim200$ hours of 720P videos for training.
To construct a balanced training set for both speaking and listening behaviors, we leverage the audio masks produced by TalkNet~\cite{talknet}, which label each video frame as either speaking or listening. For every clip, we compute the ratio of listening frames in the detected audio mask. Clips with a high listening ratio are categorized as listening data, whereas clips with a low listening ratio are categorized as speaking data. By controlling the proportion of clips selected from each category, we ensure a balanced distribution of speaking and listening samples in the final dataset.
In addition, we filter out samples with extreme head orientations by detecting the face pose and discarding videos in which the subject’s head deviates excessively from a forward-facing direction.

\paragraph{Implementation Details.}
For bidirectional teacher model training, we finetune the base model for 20000 iterations with a batch size of 32 and learning rate of 5e-6.
For the causal student model, we adopt the following set of parameters: denoising steps $N=3$, chunk size $C=3$, maximum distance for RAPR $D=9$, and the number of Q-Formers $N_Q=3$. The attention window is comprised of a 4-frame Reference Sink and a 6-frame rolling KV-cache. In Stage~1, we train the ODE initialization for 5000 iterations with a batch size of 8 and learning rate of 2e-6, followed by SiD distillation for 6000 iterations with a batch size of 16 and learning rate of 3e-6. In Stage~2, the adversarial refinement is trained for 1400 iterations with a batch size of 32 and learning rate of 5e-6. When pipelining the DiT denoising and VAE decoding on two H800 GPUs, our model achieves \emph{real-time} generation with a 1.2-second latency.

\paragraph{Evaluation Metrics.}
We comprehensively evaluate our model’s performance using multiple metrics. The Fr\'echet Inception Distance (FID)~\cite{FID} and Fr\'echet Video Distance (FVD)~\cite{FVD} assess the distributional similarity between generated and real videos. Q-Align~\cite{Q-Align} evaluates image quality (IQA) and aesthetic score (ASE), Sync-C and Sync-D~\cite{SyncScore} measure audio–lip synchronization, while Hand Keypoint Variances (HKV)~\cite{EchoMimicV2} quantify gesture dynamics.
We also adopt the Human Anomaly (HA) score from VBench-2.0~\cite{VBench-2.0} to assess distortion in body, hands, and faces.
To evaluate the motion richness during listening, we compute the variances of body, hand, and face keypoints observed during the speaker’s listening phase, denoted as LBKV, LHKV, and LFKV, respectively.

\paragraph{Compared Baselines.} 
We compare our model with state-of-the-art avatar video generation methods on both talking and interactive tasks. For talking video generation, we evaluate against Hallo3~\cite{Hallo3}, HunyuanVideo-Avatar (HY-Avatar)~\cite{HunyuanVideo-Avatar}, OmniAvatar~\cite{OmniAvatar}, EchoMimicV3~\cite{EchoMimicV3}, and StableAvatar~\cite{StableAvatar}. For interactive video generation, since no open-source full-body interactive video generation method is currently available, we implement a baseline based on OmniAvatar by feeding silent audio as the driving input during the listening phase, and compare our method against it. We also qualitatively compare with the recently proposed closed-source interactive head-and-shoulder avatar generation methods INFP~\cite{INFP} and ARIG~\cite{ARIG}, as well as the closed-source streaming interactive full-body avatar generation methods MIDAS~\cite{MIDAS} and X-Streamer~\cite{X-Streamer}.

\subsection{Qualitative Results}

\begin{figure*}
    \centering
    \includegraphics[width=0.99\linewidth]{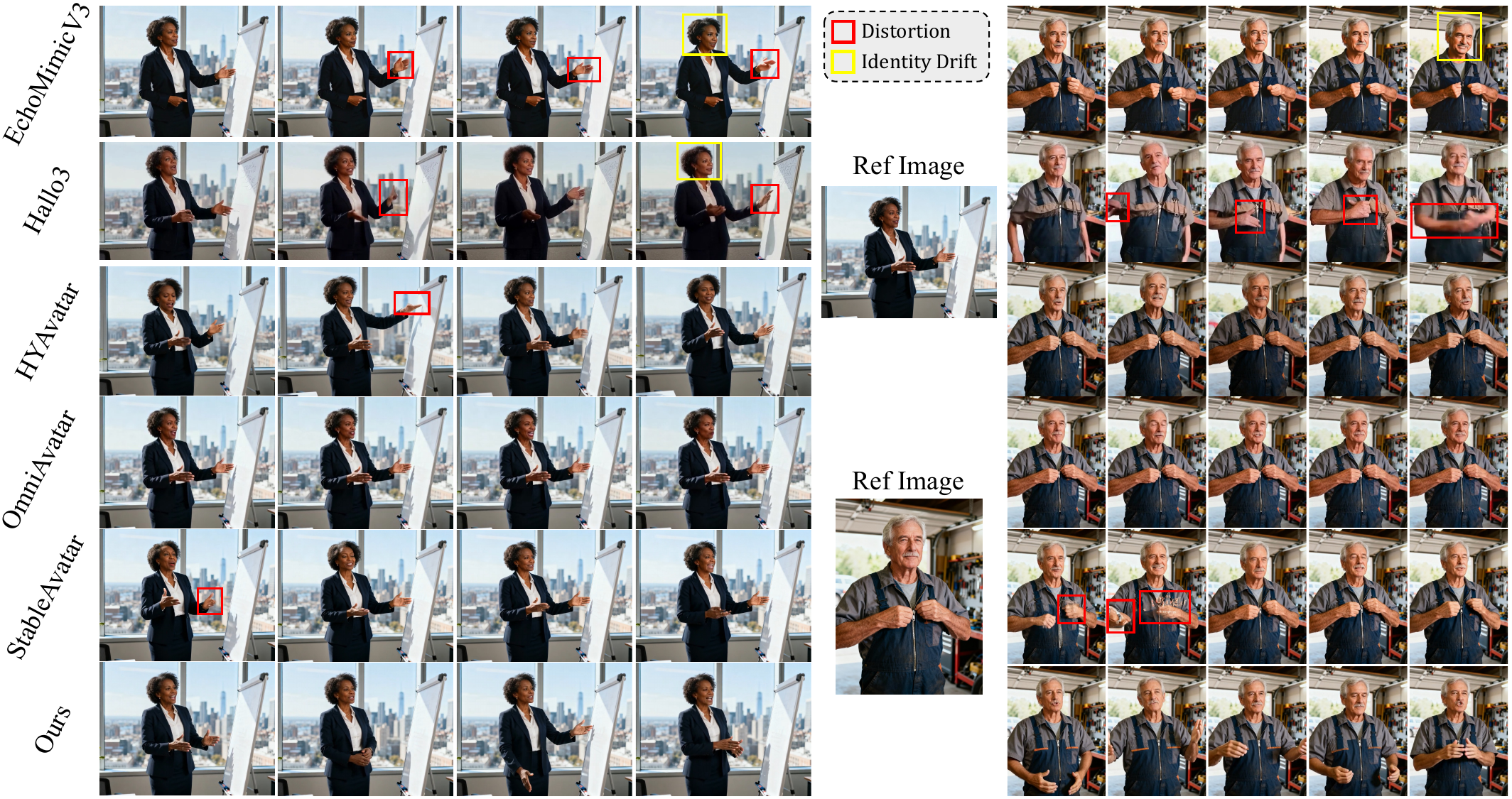}
    \caption{\textbf{Qualitative comparison with SoTA talking avatar video generation methods.} Please zoom in for details.}
    \label{fig:cmp_talk}
\end{figure*}

\begin{figure*}
    \centering
    \includegraphics[width=0.99\linewidth]{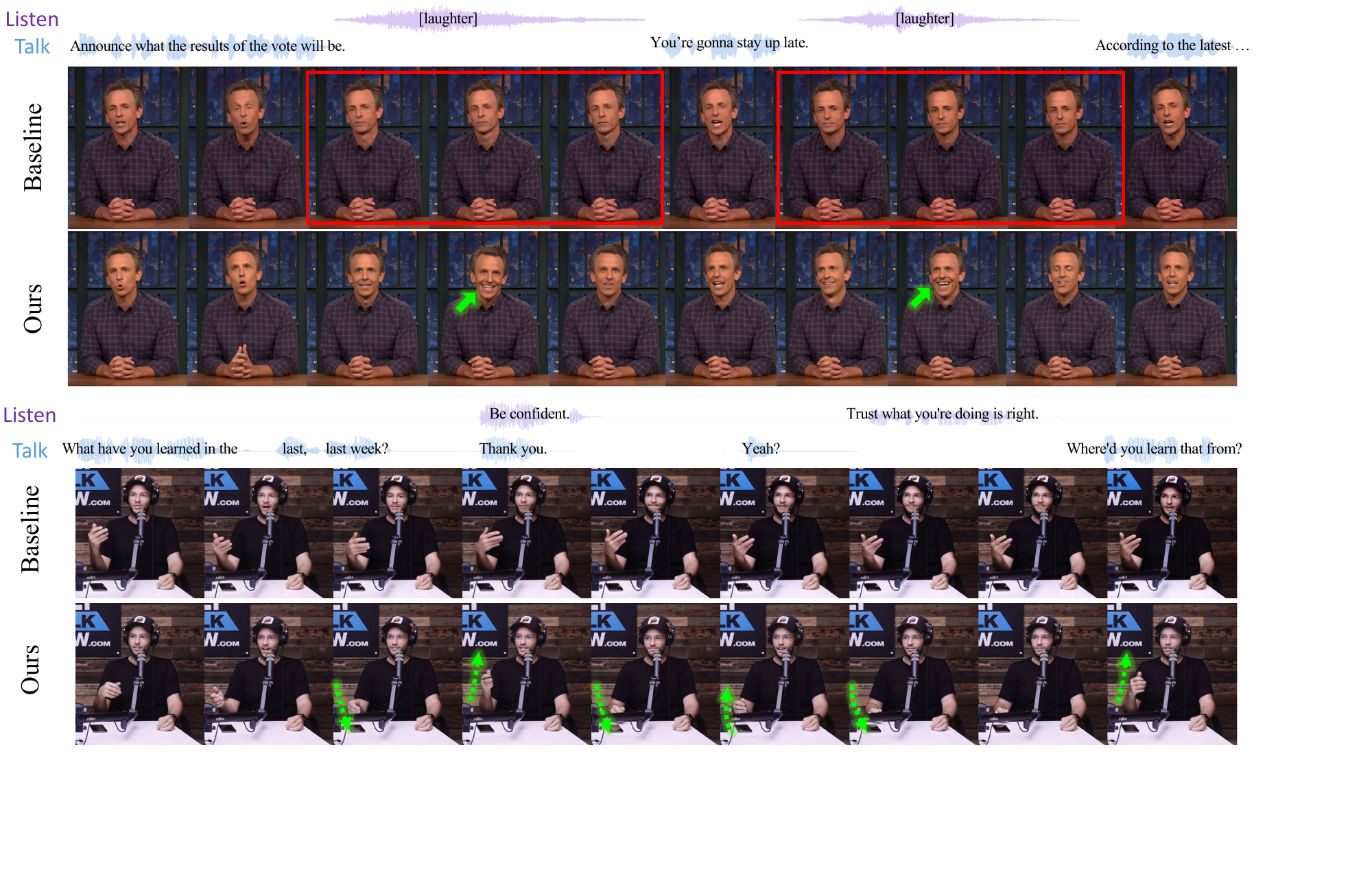}
    \caption{\textbf{Qualitative comparison with the baseline on interactive avatar video generation.} While the baseline remains almost static during the listening phase, our avatar reacts naturally to the listening audio, showing fluid and realistic transitions between talking and listening. Notice how it responds to laughter in the first example and how its expressions and gestures shift seamlessly between phases.}
    \label{fig:cmp_interact}
\end{figure*}

For talking avatar video generation, qualitative comparisons are shown in Fig.~\ref{fig:cmp_talk}. EchoMimicV3 and Hallo3 exhibit noticeable distortions, particularly in the hand regions, and suffer from error accumulation that leads to identity drift over long sequences. OmniAvatar produces stable results with fewer distortions but shows the least motion dynamics, largely preserving the posture of the reference frame. HunyuanVideo-Avatar and StableAvatar generate more dynamic motions but occasionally introduce distortions; StableAvatar further produces artifacts such as subtitles and jitters at sliding-window boundaries. In contrast, our method achieves realistic and consistent avatars with rich motion dynamics while exhibiting fewer distortions. 

For interactive avatar generation, qualitative comparisons with the baseline are shown in Fig.~\ref{fig:cmp_interact}. During the talking phase, our method produces accurate lip movements with vivid gestures, while during the listening phase, it responds naturally to auditory cues. It also generates smooth and realistic transitions between talking and listening. Compared to the baseline, which remains largely static during listening, our method exhibits greater expressiveness and realism. 

Please refer to Appendix \cref{ap_subsec:cmp_interact_head} and \ref{ap_subsec:cmp_interact_full} for additional comparisons with closed-source interactive baselines, \cref{ap_subsec:long_video} for long video generation results, and the demo video for more comparisons with all methods.

\subsection{Quantitative Results}

\begin{table*}
\centering
\caption{Quantitative comparison with SoTA talking avatar video generation methods. Metrics are reported on both short and long datasets, separated by ``/''. Best in \textbf{bold} and second best \underline{underlined}. FID/FVD for the long dataset is not applicable because the dataset is synthesized. We also report the number of denoising steps, resolution, and time taken to produce a 5-second video on a single H20 GPU for reference.}
\label{tab:cmp_talk}
\resizebox{\textwidth}{!}{%
\begin{tabular}{l|c@{\hskip 0.05in}c@{\hskip 0.1in}cccccc|c@{\hskip 0.05in}c@{\hskip 0.08in}c}
\toprule
\textbf{Method} & \textbf{FID} & \textbf{FVD} & \textbf{IQA} & \textbf{ASE} & \textbf{Sync-C} & \textbf{Sync-D} & \textbf{HKV}   &\textbf{HA}&\textbf{Steps}&\textbf{Res.}&\textbf{Time}\\
\midrule
StableAvatar & 75.20/ -& \underline{603.54}/ -& 4.66/\underline{4.87}& \underline{3.02}/3.83& 4.24/2.90& 10.92/11.57
& 42.92/38.06 &0.909/0.963
&  40&480p&12min\\
OmniAvatar & 87.24/ -& 851.93/ -& 4.45/4.66& 2.85/3.74& \textbf{7.60}/\underline{6.47}& \textbf{7.99}/8.98& 8.64/15.17
 &\textbf{0.974}/0.974&  25&480p&36min\\
HY-Avatar & 76.49/ -& \textbf{557.46}/ -& \underline{4.67}/4.80& 3.00/3.78& 6.71/5.98& 8.79/\underline{8.58}& \textbf{54.31}/\textbf{80.46}&0.947/\underline{0.975}&  50&720p&74min\\
Hallo3 & 117.41/ -& 1009.27/ -& 4.36/4.68& 2.79/3.58& 5.62/4.34& 9.70/9.61
& 35.69/59.15
 &0.958/0.959
&  51&480p&32min\\
EchoMimicV3& 78.65/ -& 724.29/ -& 4.66/\underline{4.87}& \underline{3.02}/\underline{3.90}& 3.10/2.20& 10.63/11.64& 25.53/41.31 &0.969/0.937&  25&480p&7min\\
\midrule
Ours (baseline) & 96.58/ -& 885.97/ -& 4.29/4.01& 2.75/3.13& 7.04/5.99& 8.34/8.36
&  67.65/111.69
 &0.948/0.971
&  &&\\
+ref sink & 88.75/ -& 772.10/ -& 4.55/4.64& 2.93/3.64& 7.03/5.82& 8.39/8.27
&  75.94/102.16
 &0.950/0.973
&  &&\\
+RAPR & 81.63/ -& 753.46/ -& 4.64/4.84& 2.98/3.81& 7.06/5.99& 8.34/8.18
&  48.29/61.07
 &0.956/0.961
&  &&\\
+GAN w/o  $D_{CA}$& 79.68/ -& 741.30/ -& 4.65/4.91& 3.02/3.80& 7.05/6.38& 8.31/8.15
&  35.50/29.85
 &0.947/0.993
&  &&\\
\midrule
\textbf{Ours} & \textbf{74.21}/ -& 707.34/ -& \textbf{4.68}/\textbf{4.94}& \textbf{3.03}/\textbf{3.91}& \underline{7.06}/\textbf{6.64}& \underline{8.21}/\textbf{8.14}& \underline{48.35}/\underline{65.77}&\textbf{0.974}/\textbf{0.993}& 3 &720p&20s\\
\bottomrule
\end{tabular}
} 
\end{table*}

\begin{table}
    \caption{Quantitative results for interactive avatar generation.}
    \label{tab:cmp_listen}
    \centering
    \begin{tabular}{c|ccc}\toprule
         \textbf{Method}&  \textbf{LBKV}&  \textbf{LHKV}& \textbf{LFKV}\\\midrule
         Baseline&  6.05&  4.53& 2.39\\
         Ours&  \textbf{15.88}&  \textbf{16.24}& \textbf{7.11}\\ \bottomrule
    \end{tabular}
\end{table}

For talking video generation, we construct two evaluation datasets: 1) a short dataset consisting of 50 real avatar images paired with 5-second audio clips and ground-truth videos, and 2) a long dataset consisting of 25 synthesized avatar images paired with 20-second audio clips. Quantitative results for talking avatar video generation are reported in Table~\ref{tab:cmp_talk}. Despite being only a 3-step causal model and requiring significantly less generation time, our method achieves superior or highly competitive performance across nearly all metrics, demonstrating strong visual quality and accurate lip synchronization. Notably, it achieves one of the largest motion magnitudes while maintaining a lower anomaly rate, which is challenging to accomplish simultaneously. We also report results on the EMTD~\cite{EchoMimicV2} dataset in Appendix \cref{ap_subsec:emtd}. For interactive avatar generation, we sample 50 videos from SpeakerVid-5M for evaluation. \cref{tab:cmp_listen} presents the comparison results. Our method significantly outperforms the talking-only baseline across all metrics, demonstrating its ability to produce rich motions during listening. We also conduct a user study and real-time performance analysis, with detailed results in Appendix \cref{ap_subsec:user_study} and \cref{ap_subsec:run_time}.

\subsection{Ablation Study}

We evaluate the key components of our autoregressive distillation framework and interactive model. Starting from Self Forcing~\cite{SelfForcing} as (Ours (baseline)), we incrementally add the Reference Sink (+Ref Sink), the RAPR strategy (+RAPR), adversarial refinement with a standard discriminator (+GAN w/o $D_{CA}$), and finally our consistency-aware discriminator (Ours). \cref{tab:cmp_talk} shows the quantitative results; please also refer to Appendix \cref{tab:cmp_talk_emtd}, \cref{fig:ablation}, and the demo video for additional results. The baseline fails to preserve identity and drifts over time. Introducing the Reference Sink improves identity preservation but remains insufficient for long-duration generation. The RAPR strategy further enhances temporal consistency in long sequences, though occasional blur and distortion persist. Applying adversarial refinement without the consistency-aware discriminator yields worse consistency. Note that although some ablation variants exhibit larger motion dynamics, this often comes at the cost of distorted or blurry frames, as reflected by the IQA and ASE scores. Overall, each added component brings improvement, validating our design.

To evaluate the effectiveness of our design regarding where the audio mask is applied, we introduce a ``Pre-Mask'' variant, which applies the audio mask to the raw audio before feeding it into Wav2Vec for feature extraction. In contrast, ``Ours'' applies the audio mask to the features extracted by Wav2Vec. To reduce computational cost, we conduct experiments using the undistilled model. As shown in \cref{tab:abl_listen}, ``Ours'' outperforms ``Pre-Mask'' across all metrics. This indicates that applying the audio mask directly to the audio degrades the resulting Wav2Vec features and ultimately harms model performance, thereby validating the effectiveness of our design.

\begin{table}
    \caption{Ablation on the audio mask position.}
    \label{tab:abl_listen}
    \centering
    \begin{tabular}{c|ccc}\toprule
         \textbf{Method}&  \textbf{LBKV}&  \textbf{LHKV}& \textbf{LFKV}\\\midrule
         Pre-Mask (undistilled) &  16.98&  15.49& 5.72\\
         Ours (undistilled) &  \textbf{17.74}&  \textbf{21.44}& \textbf{5.81}\\ \bottomrule
    \end{tabular}
\end{table}

\section{Conclusion}
In this paper, we address three major challenges in current human video avatars. For real-time streaming generation, we propose a two-stage autoregressive adaptation and acceleration framework with a distillation and an adversarial refinement process. For stable and consistent long video generation, we introduce Reference Sink, Reference-Anchored Positional Re-encoding, and a consistency-aware discriminator. Finally, to enable natural interaction, we develop StreamAvatar, a real-time, streaming human video model that generates both talking and listening behaviors with coherent expressions, gestures, and transitions. Extensive experiments demonstrate that StreamAvatar achieves state-of-the-art performance while operating significantly faster than competitive methods.

\paragraph{Limitations and Future Work.}
Despite its strong performance, our method has several limitations:
1) Due to the limited temporal context, it may produce inconsistent content in regions that remain occluded for an extended period. Incorporating long-term memory mechanisms could alleviate this issue.
2) While already outperforming most offline baselines in motion dynamics, the distillation process inevitably constrains the range of motion. Additionally, the current text input handling is relatively simplistic, limiting fine-grained semantic control.
Future work could leverage multimodal large language models for semantic planning to further enrich motion diversity and interactivity.
3) VAE decoding currently accounts for more than half of the total processing time. Exploring more efficient VAE decoding could further reduce streaming latency.

\section*{Acknowledgements}
\begin{sloppypar}
    This work was supported by the Natural Science Foundation of China (62461160309).
\end{sloppypar}
{
    \small
    \bibliographystyle{ieeenat_fullname}
    \bibliography{main}
}

\clearpage
\setcounter{page}{1}
\maketitlesupplementary

\section{Additional Details of the Teacher Model} \label{ap_sec:teacher}

\paragraph{Base Model Architecture.}
Our teacher model is built upon Wan2.2-TI2V-5B~\cite{Wan}, a Rectified Flow~\cite{RectifiedFlow} model comprising a causal video VAE and a bidirectional DiT denoiser. The VAE compresses video data into a compact latent space with a spatial downsampling factor of $16\times$ along both height and width, and a temporal downsampling factor of $4\times$, thereby substantially reducing the computational cost of the DiT. Concretely, a video of $n$ frames is encoded into $\lfloor(n-1)/4\rfloor + 1$ latent frames, where the first frame is encoded independently (i.e., without temporal compression) and subsequent frames are compressed at a ratio of 4. All generation and denoising operations are performed in this latent space.

The DiT takes as input a text prompt, a noisy video latent, an optional clean reference first frame, and a diffusion timestep, and predicts the Rectified Flow velocity field for denoising. Reference-image guidance is realized by replacing the first frame of the noisy video latent with the clean, noise-free latent of the reference image; the bidirectional self-attention mechanism then propagates identity and appearance information from this anchor frame to all subsequent frames and achieve spatial and temporal coherence. Cross-attention layers inject the text-prompt information into the video latent to enable text-based control.

\paragraph{Audio Encoder.}
To obtain audio features suitable for injection into the video model as driving conditions, we design an audio encoder as illustrated in the yellow region of \cref{fig:structure}.
We extract multi-layer deep features from a pretrained Wav2Vec 2.0~\cite{Wav2Vec2} encoder. Because the VAE's temporal compression ratio differs between the first frame and subsequent frames, an explicit step is required to align Wav2Vec features with VAE latent frames.

We adopt a context-window approach: each latent frame attends to a short temporal neighborhood of Wav2Vec features centered around its corresponding video frame, so that anticipatory and carry-over acoustic cues (e.g., mouth opening before speech onset, or a prolonged sigh) are captured. Denoting the Wav2Vec feature corresponding to the uncompressed video frame $i$ as $f_i$ (with $f_i = f_0$ for $i < 0$), the audio feature $f'_t$ assigned to latent frame $t$ is defined as:
\begin{equation} \label{eq:audio_extraction}
    f'_t = 
    \begin{cases}
        \mathrm{concat}\!\left(\{f_i\}_{i=t-2}^{t+2}\right), & t = 0, \\[4pt]
        \mathrm{concat}\!\left(\{f_i\}_{i=4t-5}^{4t+2}\right), & t > 0.
    \end{cases}
\end{equation}
Because $f'_0$ aggregates 5 Wav2Vec frames while $f'_t\;(t>0)$ aggregates 8, their raw dimensions differ. We therefore apply separate lightweight MLP projectors (denoted ``Audio Proj.'') to map both cases to a common feature dimension. Importantly, the talking and listening streams use \emph{independent} projectors, allowing each to learn phase-specific representations. After projection, the features are concatenated along the temporal axis to form frame-aligned audio feature sequences, yielding the talking audio feature sequence $\{a_{\text{talk},t}\}$ or the listening audio feature sequence $\{a_{\text{listen},t}\}$.

\paragraph{Audio Attention Modules.}
As shown in the cyan region of \cref{fig:structure}, to enable speech-driven generation of both talking and listening motions, we insert audio attention modules into each of the $N_\text{blk}=30$ DiT blocks of the video model, injecting audio information via cross-attention between the video latents and the audio features.
To preserve strict temporal correspondence between audio and motion, each latent frame's query tokens attend \emph{only} to the audio features assigned to that same frame, rather than to the full audio sequence. This frame-wise cross-attention design prevents temporal leakage and ensures that lip movements and gestures remain tightly synchronized with the driving audio. All other layers in the Transformer block (self-attention, text cross-attention, and feed-forward layers) remain unmodified and audio-agnostic.

\paragraph{Training Procedure.}
We train the teacher model following the standard Rectified Flow training paradigm under the Flow Matching~\cite{FlowMatching} framework. Given a video latent $x^0$ and its corresponding conditions (reference first frame, audio features, and text prompt), we sample a random timestep $n \in (0,1)$ and construct a noisy latent $x^n$ by linearly interpolating between $x^0$ and Gaussian noise $\epsilon$: $x^n = (1-n)x^0 + n\epsilon$. The model is trained with a mean squared error loss to predict the velocity field $v = \epsilon - x^0$ at the sampled point. During training, we adopt a two-stage strategy: we first freeze the pretrained DiT and only train the audio projection and audio attention modules, then unfreeze the DiT and fine-tune all parameters jointly.

\section{Additional Experiments}

\subsection{Quantitative Comparison/Ablation on EMTD} \label{ap_subsec:emtd}

To further evaluate our approach, we compare it with baseline methods and ablation variants on the EchoMimicV2 Testing Dataset (EMTD)~\cite{EchoMimicV2}. The EMTD dataset contains 110 front-facing, half-body speech videos. Quantitative results are presented in \cref{tab:cmp_talk_emtd}. Our method outperforms all comparison methods across almost all metrics, and the ablation results further demonstrate the effectiveness of our design.

\begin{table*}[h]
\centering
\caption{Quantitative comparison with SoTA talking avatar video generation methods on the EMTD dataset. Best in \textbf{bold} and second best \underline{underlined}.}
\label{tab:cmp_talk_emtd}
\begin{tabular}{l|cccccccc}
\toprule
\textbf{Method} & \textbf{FID} & \textbf{FVD} & \textbf{IQA} & \textbf{ASE} & \textbf{Sync-C} & \textbf{Sync-D} & \textbf{HKV}  &\textbf{HA}\\
\midrule
StableAvatar & 91.63 & 840.86 & 3.67 & 2.37 & 3.04 & 12.16 
& 57.99 
&  0.794 \\
OmniAvatar & 75.20 & 982.09 & 3.72 & 2.45 & \underline{7.68}& \underline{7.97}& 29.04 
&  0.889 \\
HY-Avatar & \underline{63.09}& \underline{765.05}& 3.91 & 2.57 & 7.35 & 8.36 
& \textbf{66.07}&  0.880 \\
Hallo3 & 91.15 & 898.19 & 3.57 & 2.26 & 5.62 & 9.72 
& 29.52 
&  0.874 \\
EchoMimicV3& 67.35 & 822.89 & \underline{3.98}& \underline{2.68}& 3.00 & 12.20 & 56.58 &  \underline{0.921}\\
\midrule
Ours (baseline) & 107.50 & 1254.66 & 3.25 & 2.12 & 7.23 & 8.26 
&  80.15 
&  0.934 \\
+ref sink & 81.10 & 1060.86 & 3.69 & 2.36 & 7.60 & 8.05 
&  79.67 
&  0.908 \\
+RAPR & 63.71 & 801.68 & 4.03 & 2.66 & 7.45 & 8.04 
&  62.09 
&  0.925 \\
+GAN w/o $D_{CA}$ & 59.87 & 749.32 & 4.03 & 2.64 & 7.67 & 7.88 
&  36.79 
&  0.929 \\
\midrule
\textbf{Ours} & \textbf{61.84}& \textbf{683.14}& \textbf{4.13}& \textbf{2.78}& \textbf{8.06}& \textbf{7.93}& \underline{62.60}& \textbf{0.935}\\
\bottomrule
\end{tabular}
\end{table*}

\begin{figure*}[h]
  \centering
   \includegraphics[width=\linewidth]{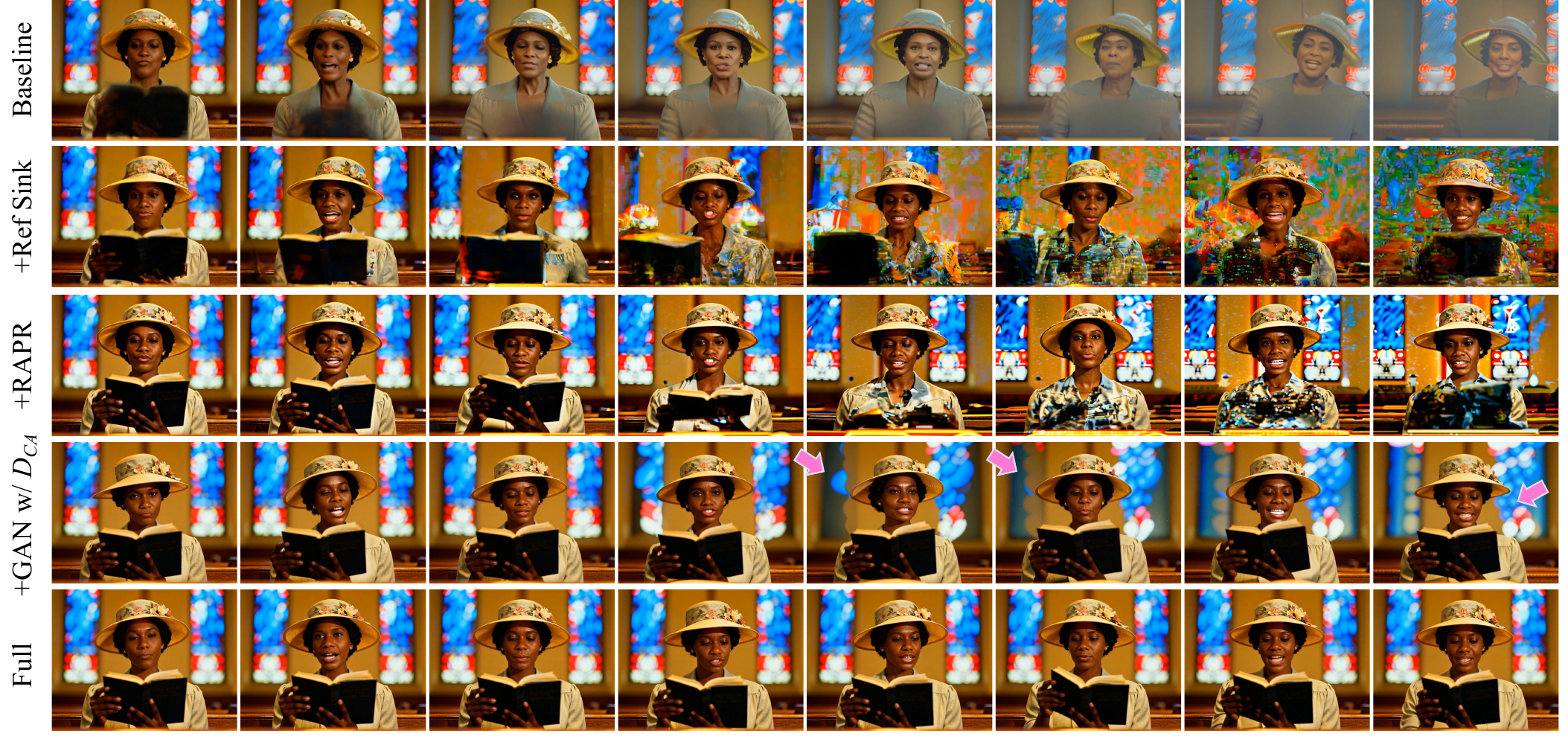}
   \caption{Qualitative Ablation Results.}
   \label{fig:ablation}
\end{figure*}

\subsection{Qualitative Ablation Results}

We present qualitative ablation results in \cref{fig:ablation} and the demo video. Note how the addition of key components gradually improves long-term generation stability and consistency, identity preserving, and visual quality.

\subsection{User Study} \label{ap_subsec:user_study}

We conduct a user study to comprehensively evaluate our method. Participants are shown paired video clips generated by our approach and a comparison method, and asked to assess the two along five dimensions: audio–lip synchronization (Sync), motion dynamics (Dynamics), temporal continuity and smoothness (Continuity), visual quality and naturalness (Quality), and identity preservation (Identity). For each pair, participants indicate whether they prefer our method, prefer the comparison method, or have no preference. In total, we collect 960 paired comparisons from 24 participants. As illustrated in \cref{tab:user_study}, our method consistently outperforms the state-of-the-art baselines across almost all comparisons, which aligns closely with our quantitative evaluation.

\begin{table*}[!h]
\centering
\caption{\textbf{User study results.} The table presents the pairwise preference rates (\%) across different metrics, formatted as (Ours / Baseline). Winning values are highlighted in \textbf{bold}. The remaining percentage in each comparison accounts for ``Tie'' (no preference) cases.}
\label{tab:user_study}
\begin{tabular}{l|ccccc}
\toprule
\textbf{Ours vs X} & \textbf{Sync (\%)} & \textbf{Quality (\%)} & \textbf{Dynamics (\%)} & \textbf{Identity (\%)} & \textbf{Continuity (\%)} \\
\midrule
EchoMimicV3   & \textbf{91.4} / 2.1 & \textbf{68.6} / 4.9 & \textbf{74.1} / 8.6 & \textbf{47.6} / 1.0 & \textbf{50.8} / 1.1 \\
Hallo3        & \textbf{86.2} / 2.7 & \textbf{79.9} / 2.6 & \textbf{47.1} / 28.6 & \textbf{64.6} / 1.0 & \textbf{68.3} / 1.5 \\
HY-Avatar     & \textbf{41.2} / 18.0 & \textbf{48.5} / 13.4 & 16.5 / \textbf{57.2} & \textbf{28.9} / 7.7 & \textbf{44.3} / 8.8 \\
OmniAvatar    & \textbf{45.9} / 14.8 & \textbf{53.6} / 6.1 & \textbf{75.0} / 13.3 & \textbf{25.5} / 4.1 & \textbf{28.1} / 6.1 \\
StableAvatar  & \textbf{74.0} / 5.6 & \textbf{65.3} / 5.1 & \textbf{39.3} / 36.7 & \textbf{61.7} / 3.6 & \textbf{66.8} / 4.6 \\
\bottomrule
\end{tabular}
\end{table*}

\begin{figure*}[!h]
  \centering
  \begin{minipage}[t]{0.48\textwidth}
    \centering
    \includegraphics[width=\linewidth]{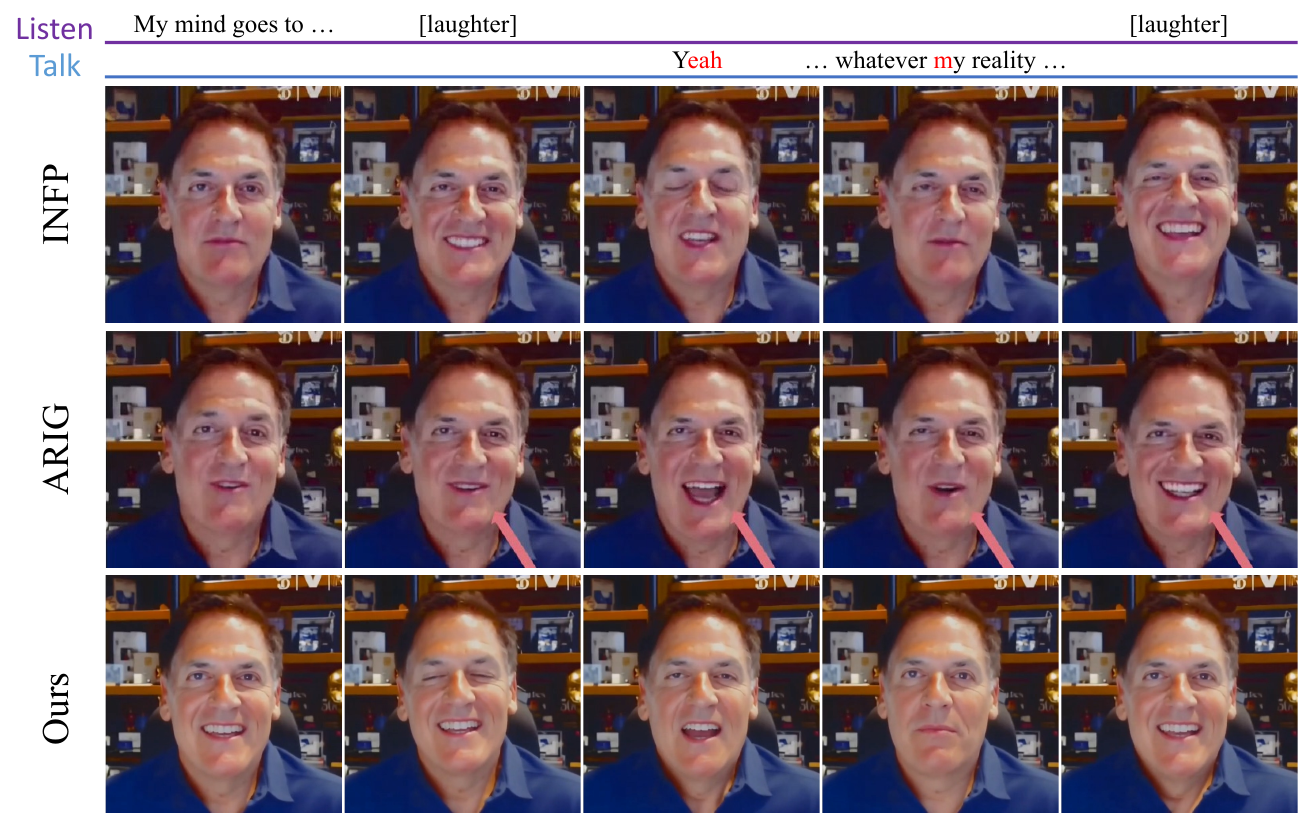}
    \caption{Qualitative comparison with SoTA interactive head generation methods. Please ignore the arrows which come with the original video on ARIG's project page.}
    \label{fig:cmp_head}
  \end{minipage}
  \hfill
  \begin{minipage}[t]{0.48\textwidth}
    \centering
    \includegraphics[width=\linewidth]{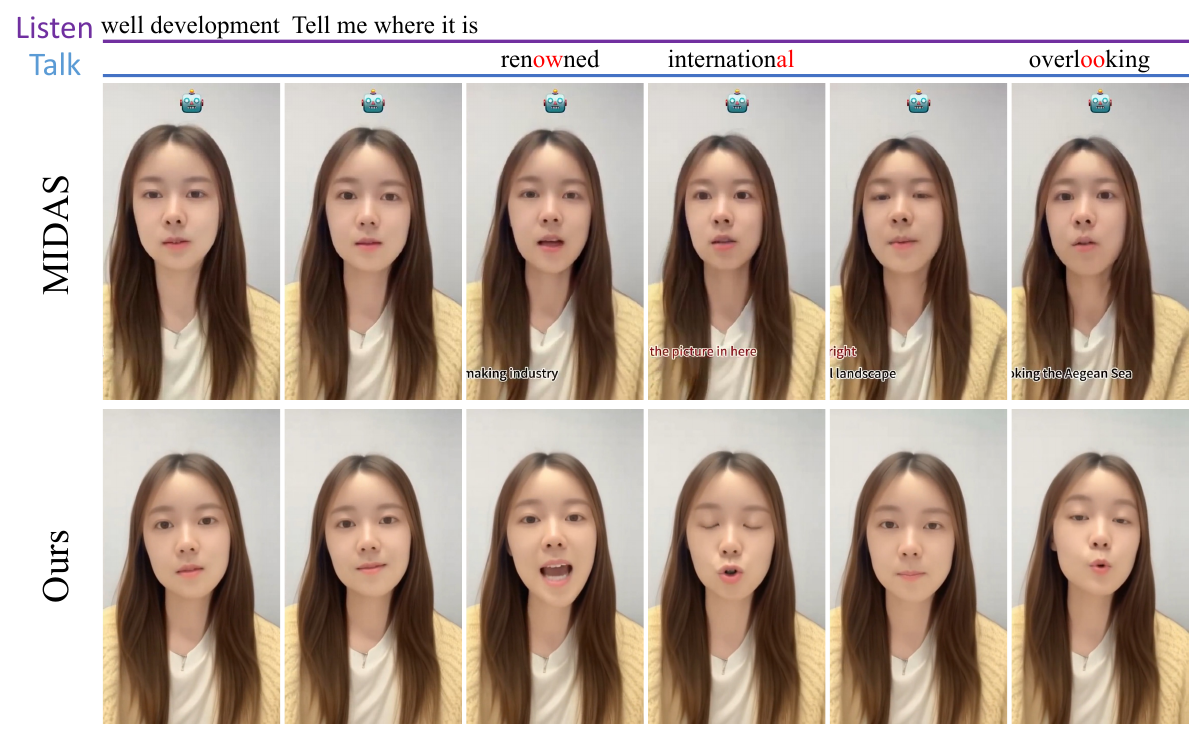}
    \caption{Qualitative comparison with MIDAS~\cite{MIDAS}.}
    \label{fig:cmp_midas}
  \end{minipage}
\end{figure*}

\begin{figure*}[!h]
  \centering
   \includegraphics[width=0.8\linewidth]{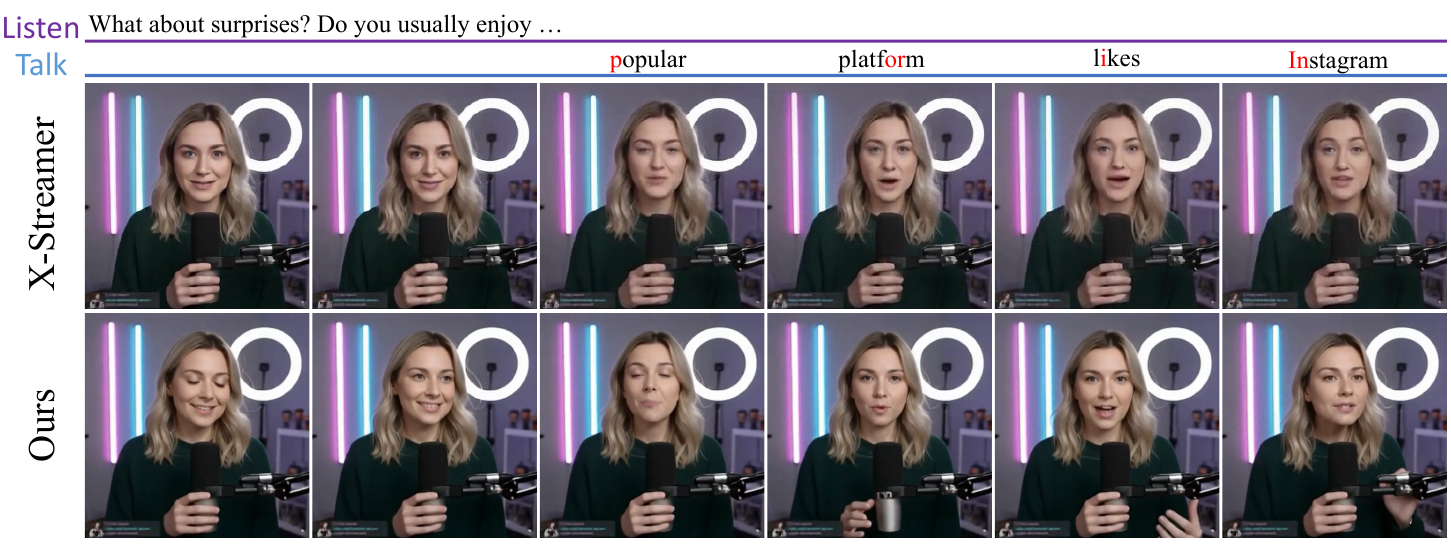}
   \caption{Qualitative comparison with X-Streamer~\cite{X-Streamer}.}
   \label{fig:cmp_xstreamer}
\end{figure*}

\subsection{Comparison with Interactive Head Generation} \label{ap_subsec:cmp_interact_head}

We compare our method with state-of-the-art interactive head generation methods, including INFP~\cite{INFP} and ARIG~\cite{ARIG}. As these methods are not open-sourced, we conduct qualitative comparison with the results from their project pages, as shown in \cref{fig:cmp_head} and the demo video. Although our model is designed and trained for body video generation, it still performs on par with these dedicated head avatar methods, while delivering the best visual quality.

\subsection{Comparison with Streaming Interactive Avatar Generation} \label{ap_subsec:cmp_interact_full}

We further compare our method with current state-of-the-art streaming interactive avatar generation methods, including MIDAS~\cite{MIDAS} and X-Streamer~\cite{X-Streamer}. As these methods are not open-sourced, we conduct qualitative comparison with the results from their project pages, as shown in \cref{fig:cmp_midas}, \cref{fig:cmp_xstreamer}, and the demo video. Our method produces more accurate lip synchronization and more vivid expressions than MIDAS. It is also worth noting that our method is \emph{one-shot}, whereas MIDAS requires person-specific finetuning. Our method also exhibits more natural listening behaviors, more diverse motions, and higher visual quality than X-Streamer.

\subsection{Long Video Generation} \label{ap_subsec:long_video}

\begin{figure*}[t]
   \centering
   \includegraphics[width=1\linewidth]{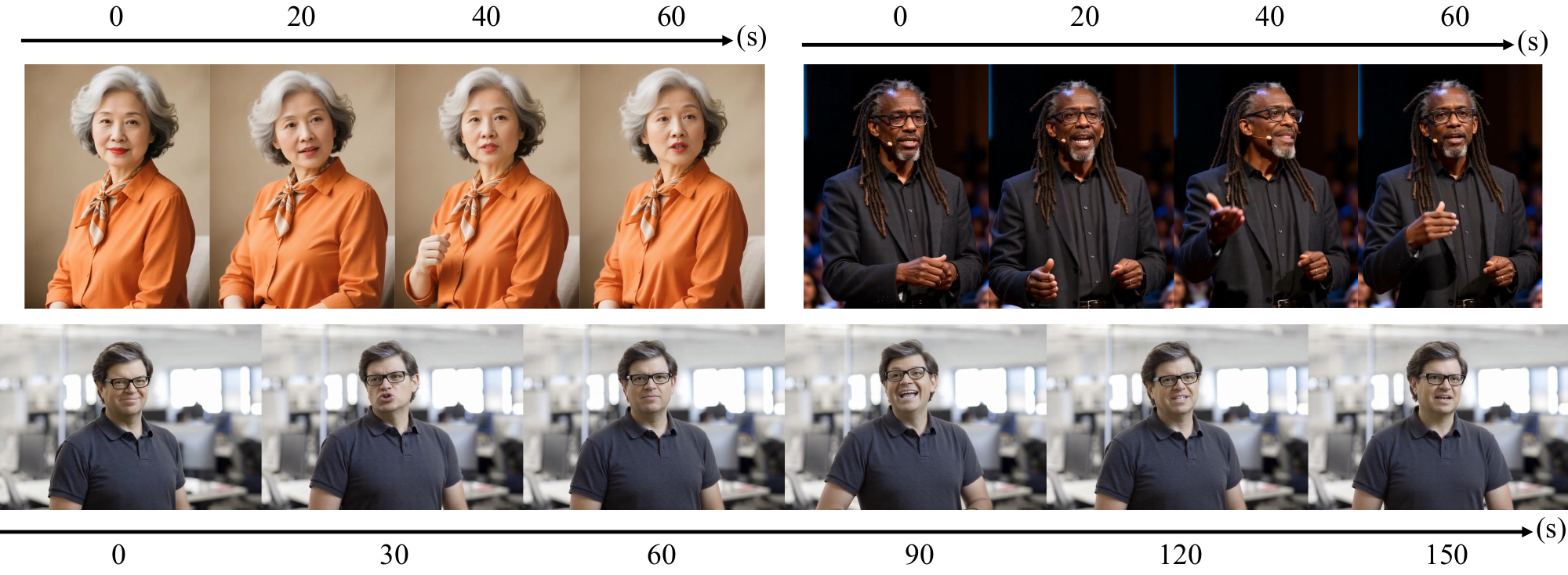}
   \caption{Long video generation results.}
   \label{fig:long_video}
\end{figure*}

Thanks to the streaming architecture with all our proposed techniques to improve consistency and stability for long video generation, our approach can generate videos of arbitrary length without quality degradation in real-time. \cref{fig:long_video} shows sampled frames from generated videos of up to 150 seconds. The results maintain consistent identity, appearance, and visual quality throughout the entire duration. We refer readers to the supplementary demo video for the full-length results.

\section{Runtime Details} \label{ap_subsec:run_time}

\begin{table}[!h]
    \caption{Evaluation of the real-time performance of our model.}
    \label{tab:real_time}
    \centering
    \begin{tabular}{c|cc}\toprule
         Module&  RTF& FFD\\\midrule
         DiT&  0.69& 0.33s\\
         VAE&  0.82& 0.39s\\ \bottomrule
    \end{tabular}
\end{table}

Our model generates videos at 25 FPS. To enable real-time generation, we distribute the DiT denoising and VAE decoding processes across two NVIDIA H800 GPUs. We evaluate the performance under our default model configuration (denoising steps $N{=}3$, chunk size $C{=}3$, total KV cache length $L{=}10$) when generating videos at a resolution of $928\times704$. We report two metrics: the Real-Time Factor (RTF), defined as the ratio between the inference time and the duration of the generated video segment, and the First Frame Delay (FFD), defined as the time elapsed from receiving the input to producing the first output frame.

The results are listed in \cref{tab:real_time}. Since the RTF values of all modules are below 1, the system supports real-time generation.

\paragraph{Audio Lookahead.}
Recall from the audio feature extraction in \cref{eq:audio_extraction} that the context window for latent frame $t>0$ includes two ``future'' Wav2Vec features $f_{4t+1}$ and $f_{4t+2}$, which correspond to video frames that have not yet been observed at generation time. We empirically find that replacing these two features with $f_{4t}$ (i.e., repeating the current-frame feature) has a negligible effect on generation quality. Consequently, in the deployed system the model does not need to wait for any future audio input beyond the current chunk boundary. The overall system latency is therefore given by the sum of the FFD and the input chunk buffering delay ($C\times 4/25 = 0.48$s), yielding a total end-to-end latency of approximately 1.20s.

\begin{figure}[!h]
  \centering
   \includegraphics[width=\linewidth]{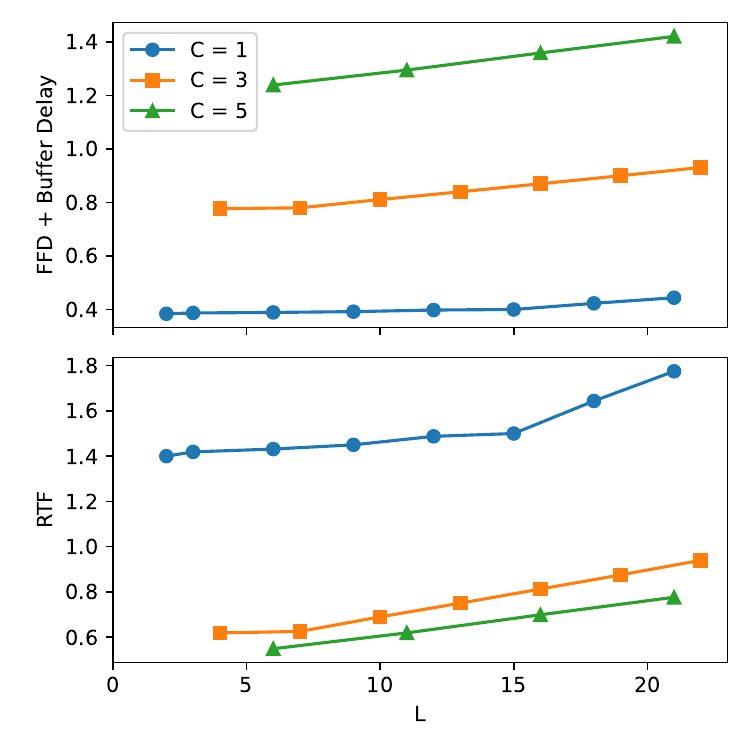}
   \caption{Impact of chunk size $C$ and KV cache total length $L$ on real-time performance. \textbf{Top:} First Frame Delay (FFD). \textbf{Bottom:} Real-Time Factor (RTF).}
   \label{fig:real_time}
\end{figure}

\paragraph{Impact of Chunk Size and KV Cache Length.}
We further analyze how the chunk size $C$ and the total KV cache length $L$ affect real-time performance, as illustrated in \cref{fig:real_time}. A smaller $C$ reduces the first-frame delay because each chunk covers fewer frames; however, it also increases the per-frame overhead and decreases the intra-chunk parallelism, pushing RTF above 1 and breaking the real-time constraint. Conversely, a larger $C$ improves throughput (lower RTF) at the cost of higher latency. Increasing $L$ provides the model with a longer temporal context, which benefits temporal consistency, but also raises both latency and RTF because each attention operation must attend to more cached tokens. Too small an $L$, on the other hand, degrades temporal coherence as the model loses access to sufficient history. Our default configuration ($C{=}3$, $L{=}10$) strikes a balanced trade-off among latency, throughput, and generation quality.

\section{Ethical Considerations}

This work focuses on talking avatar generation for constructive, human-centered applications, and is not intended to support deceptive or harmful media. As with any generative technology, misuse is possible, such as creating fraudulent identities, fabricating false narratives, or generating avatars for harassment. To mitigate these risks, we commit to safeguards including embedding watermarks and clearly disclosing that all outputs are synthetic when deploying the technology. We also aim to collaborate with the research community to develop improved deepfake detection tools and support efforts to establish standards for media provenance.

\end{document}